\pdfoutput=1

\documentclass[11pt]{article}
\usepackage{acl}

\usepackage{times}
\usepackage{latexsym}

\usepackage[T1]{fontenc}

\usepackage[utf8]{inputenc}

\usepackage{microtype}

\usepackage{graphicx} 
\usepackage{float} 
\usepackage{stfloats}
\usepackage{subfigure} 
\usepackage{arydshln}
\usepackage{pgfplots}
\usepackage{pgfplotstable}

\usepackage{booktabs}
\usepackage{siunitx}
\usepackage{multirow}
\usepackage{makecell} 
\usepackage[table,xcdraw]{xcolor}
\usepackage{dashrule}
\usepackage{arydshln}
\usepackage{array}
\usepackage{booktabs}

\usepackage{mathrsfs}
\usepackage{amsmath}
\usepackage{amssymb}

\usepackage{tcolorbox}

\title{Track-SQL: Enhancing Generative Language Models with Dual-Extractive Modules for Schema and Context Tracking in Multi-turn Text-to-SQL}

\author{
    Bingfeng Chen$^{1,2}$, Shaobin Shi$^{1}$, Yongqi Luo$^{1}$, Boyan Xu$^{1}$\thanks{Corresponding author, \href{mailto:hpakyim@gmail.com}{hpakyim@gmail.com}}, Ruichu Cai$^{1,3}$, Zhifeng Hao$^{1,4}$ \\
    $^{1}$School of Computer Science, Guangdong University of Technology \\
    $^{2}$Guangdong Laboratory of Artificial Intelligence and Digital Economy (SZ) \\
    $^{3}$Peng Cheng Laboratory \\
    $^{4}$College of Science, Shantou University \\
    \texttt{chenbf@gdut.edu.cn} \\
    \texttt{\{d7inshi, lyongqi001, hpakyim, cairuichu\}@gmail.com} \\
    \texttt{haozhifeng@stu.edu.cn} \\
}

\begin{document}
\maketitle

\begin{abstract}
Generative language models have shown significant potential in single-turn Text-to-SQL. However, their performance does not extend equivalently to multi-turn Text-to-SQL. This is primarily due to generative language models' inadequacy in handling the complexities of context information and dynamic schema linking in multi-turn interactions. In this paper, we propose a framework named Track-SQL, which enhances generative language models with dual-extractive modules designed to track schema and contextual changes in multi-turn Text-to-SQL. Specifically, Track-SQL incorporates a \emph{Semantic-enhanced Schema Extractor} and a \emph{Schema-aware Context Extractor}. Experimental results demonstrate that Track-SQL achieves state-of-the-art performance on the SparC and CoSQL datasets. Furthermore, detailed ablation studies reveal that Track-SQL significantly improves execution accuracy in multi-turn interactions by 7.1\% and 9.55\% on these datasets, respectively. Our implementation will be open-sourced at \url{https://github.com/DMIRLAB-Group/Track-SQL}.
\end{abstract}

\section{Introduction}
Text-to-SQL \cite{zhong2017seq2sqlgeneratingstructuredqueries} is a critical semantic parsing task that converts natural language queries into corresponding SQL statements based on a given database schema.  
While generative language models have demonstrated significant potential in single-turn Text-to-SQL tasks, their performance diminishes in multi-turn scenarios. The challenges in these settings primarily stem from the models' difficulties in handling the complexities of context information and dynamic schema linking across multiple turns of interaction. However, extractive model to solving these two challenges—schema linking and context utilization—have limitations when directly applied to the generative language model paradigm.

The primary limitation in multi-turn Text-to-SQL is the ability to maintain effective schema linking as the dialogue progresses. Although many studies have confirmed that proper schema linking significantly enhances SQL generation, existing approaches struggle to handle the increasing complexity in multi-turn settings. 
In recent research, RASAT \cite{qi-etal-2022-rasat} leverages relational self-attention mechanisms to capture the relationships between text and schemas. However, in multi-turn dialogue scenarios, as the number of interactions between the user and the system increases along with the expansion of the database schemas, the scale of the schema linking graph grows, inevitably leading to the issue of redundant links. CQR-SQL \cite{xiao2022cqr} rewrites multi-turn dialogues and simplifies schema linking information, which might also result in the loss of critical schema linking details due to over-simplification. TP-Link \cite{liu-etal-2024-tp} integrates schema linking prediction into multi-task pre-training to reduce redundant relationships between questions and schemas, but overlooks the problem of semantic inconsistencies between questions and schemas. 

Moreover, existing schema linking methods are predominantly static, lacking mechanisms to incorporate linking results from prior turns. The key idea to overcome this limitation is to introduce a dynamic updating mechanism that adapts to the evolving dialogue contexts in multi-turn interactions.

The secondary limitation in multi-turn Text-to-SQL is their difficulty in managing continuous interactions where users reference or omit prior information, relying on the system to track the evolving context. In recent research, 

EditSQL \cite{zhang-etal-2019-editing} uses prior turn SQL queries for predicting current turn queries, but loses effectiveness when dialogue lacks coherence. 
STaR \cite{cai-etal-2022-star} and CoE-SQL \cite{zhang2024coesqlincontextlearningmultiturn} track dependencies via SQL similarity and changes, but lack source record verification, which can lead to error buildup. The key idea to overcome this limitation is to design effective retrieval and verification mechanisms for improving the accuracy and reliability of multi-turn Text-to-SQL.

To address these issues, we propose a Track-SQL framework, aimed at solving the problems of dynamic schema linking and context information filtering in multi-turn Text-to-SQL dialogues. First, in the schema extraction phase, we developed a \emph{Semantic-enhanced Schema Extractor} (SESE) that identifies the user's current focus schemas by combining changes in user interests with previously extracted signals. At the same time, we introduce a semantic enhancement module to reduce the semantic gap between user questions and schemas, thereby improving the precision of schema linking. Second, in the SQL generation phase, we design a \emph{Schema-aware Context Extractor} (SACE) module to identify key SQLs from historical records.
Then, we combine these with past questions and extracted schemas as input, and fine-tune the text-to-SQL generation model in a supervised manner, with the target output being normalized SQL queries.
This approach reduces the difficulty for the generation model to learn schema linking and context information filtering, thus enhancing the accuracy of SQL generation. Additionally, the results from the aforementioned extractors make the basis for the model's SQL generation more transparent, increasing the explainability of the system.

Our contributions are summarized as follows:
\begin{itemize}
\item  We propose the Track-SQL framework, specifically designed for Multi-turn Text-to-SQL tasks. This framework utilizes a Semantic-enhanced Schema Extractor to ensure that the SQL generation model acquires accurate schema information in each dialogue turn.
\item We have devised a Schema-aware Context Extractor to obtain the most relevant historical SQL queries that fit the current conversational context, thereby enhancing the dialogue history understanding capability and SQL generation accuracy of the generative language model.
\item We conducted extensive performance evaluations and detailed ablation studies to verify the effectiveness of each component. The Track-SQL framework achieved leading results on the validation sets of two authoritative benchmark datasets, SParC and CoSQL, demonstrating its superior performance and broad applicability.
\end{itemize}

\section{Methodology}
\begin{figure*}[h]
    \centering
    \makebox[\textwidth][c]{
        \includegraphics[width=1\linewidth]{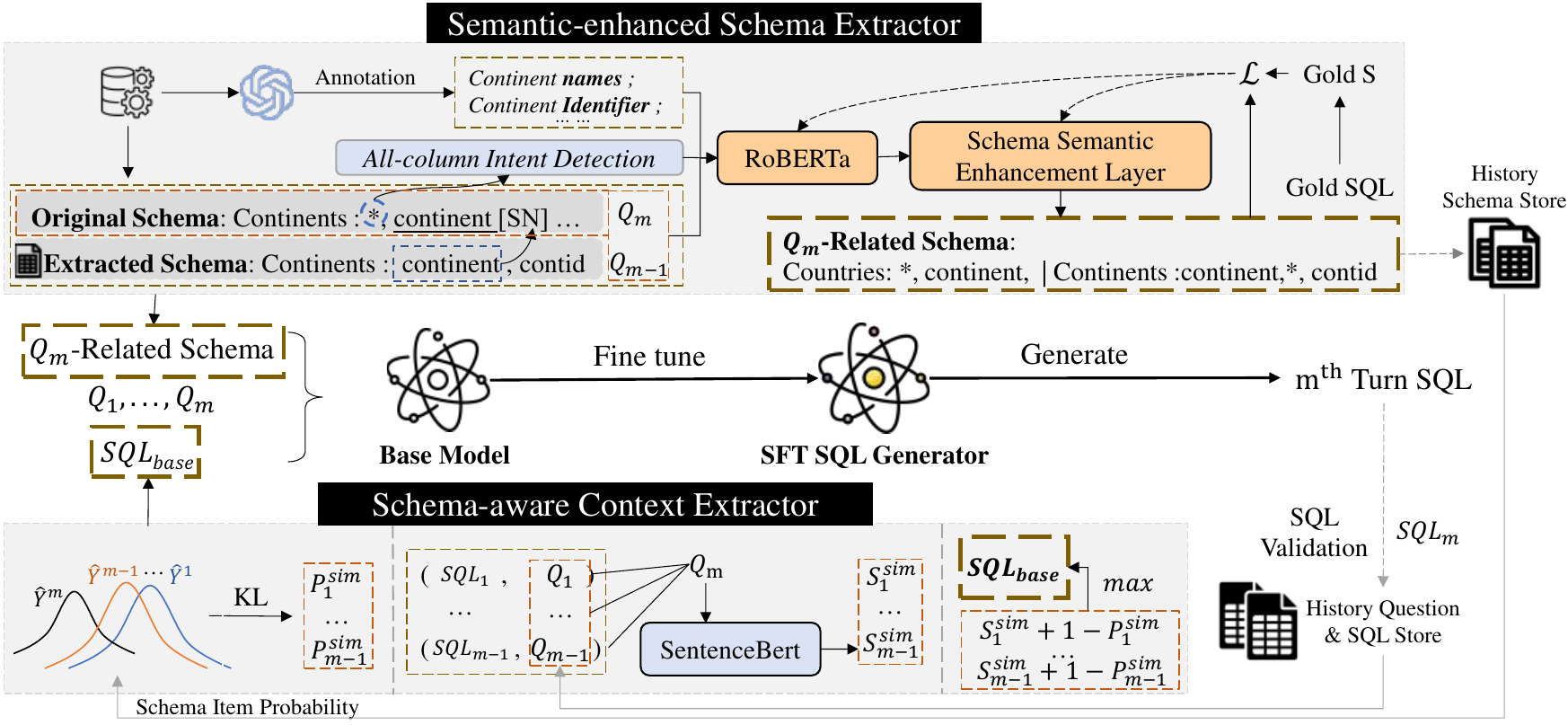}
    }
    \caption{The overall framework of Dual-Extractive Modules for Schema and Context Tracking. The framework trains a schema item classification model and an SQL generator. Based on the former, we construct a \emph{Semantic-enhanced Schema Extractor} and a \emph{Schema-aware Context Extractor}. The extraction results from these two extractors are utilized for subsequent training of the SQL generation model. The core idea of Track-SQL is to reduce the gap between the input and the target SQL before entering the multi-turn SQL generation phase by means of dynamic schema linking and context information extraction.}
    \label{model}
\end{figure*}

In the multi-turn Text-to-SQL task, the goal is to address the problem of mapping a sequence of multi-turn questions $\mathcal{Q}_{\leq m}$ and database schemas $\mathcal{S}=(t_{i},c_{i,n_{i}})$ to the target SQL query $s_{m}$, where $t_{i}$ represents the $i^{th}$ table in the database, and $c_{i,n_{i}}$ denotes the $n_{i}^{th}$ column within $t_{i}$. This section will provide an overview of the framework designed to solve this problem and delve into its design details.

\subsection{Model Overview}
In multi-turn Text-to-SQL tasks, we decompose two preparatory tasks: Dynamic Schema Linking and Context Information Filtering. Enhancing the performance of the generative language model is achieved through utilizing dual-extractive modules to identify key schema and contextual information. Figure \ref{model} illustrates the overall architecture of the proposed Track-SQL framework, which includes two history information repositories and their corresponding extraction modules, along with a supervised fine-tuned SQL generator. Specifically, during the $m^{th}$ interaction, we input the first $m$ questions and all schema information into Schema Extractor to obtain the probabilities of all schemas for the current question $\mathcal{Q}_{m}$. Based on these probabilities and using fixed threshold $s$, we filter and rank the schemas (Section \ref{sec:Multi-turn Schema Item Extractor}). Additionally, by utilizing the schema probabilities stored in the History Schema Store, we can resolve the coreference relationships between $\mathcal{Q}_{m}$ and $\mathcal{Q}_{< m}$ and select the base SQL from History Question$\&$SQL Store
which applicable to current question (Section \ref{sec:History SQL Selection}). These filtered schemas and base SQL serve together as prompt information and constraints to facilitate the generation of SQL at the $m^{th}$ turn.

\subsection{Semantic-enhanced Schema Extractor}
\label{sec:Multi-turn Schema Item Extractor}
In the context of multi-turn dialogues involving databases, redundant schema item information can significantly interfere with the generation of SQL queries. To address this issue, we designed SESE to filter out redundant table column information. 
The extractor consists of three intertwined sub-elements: Historical Extraction Item Tagging, Schema Semantic Enhancement and All-Column Intent Detection. These elements respectively facilitate dynamic schema linking encoding, semantic alignment between question entities and schemas, and all-column intent encoding.

Regarding the specific implementation details, we first define the original sequence of concatenated multi-turn questions and schemas as: $\mathcal{X} = \mathcal{Q}_{1} \ \& \  ... \ \&\ \mathcal{Q}_{m}\ |\  t_{1} : c_{11}, ... , c_{1n_{1} } \ |\  ... \ |\  t_{N} : c_{N1} , ... ,c_{Nn_{N} }$, where $\&$ connects multiple turns of questions, and $|$ separates different schemas. To enhance the semantic expressiveness of the schemas, we introduce open-domain semantic knowledge, combining database content with large language models (LLMs) to enrich the semantic information of column names, and further utilize the enhanced column names to enrich the semantic information of table names. Specifically, we sample some values randomly from each column in each table, along with the type and name of the column, as inputs to prompt LLM to generate descriptive comments; then, based on the information of all columns, generate comments for the tables. Thus, we obtain a schema annotation sequence: $\hat{\mathcal{S}} = \hat{t}_{1} : \hat{c}_{11}, ... , \hat{c}_{1n_{1} } \ |\  ... \ |\ \hat{t}_{N} : \hat{c}_{N1} , ... ,\hat{c}_{Nn_{N} }$. In a multi-turn interactive environment, we retrieve the columns extracted from the previous turn from the history schema store and mark these columns using the symbol $\mathbf{[SN]}$ within the current turn's input $\mathcal{X}$. Subsequently, $\mathcal{X}$ and $\hat{S}$ are sequentially input into RoBERTa \cite{liu2019robertarobustlyoptimizedbert}. To integrate and classify the schemas along with their corresponding comments as a complete unit, we perform pooling operations on the output embeddings of each token after tokenization by RoBERTa for the schemas. To accomplish this, we used a pooling module composed of two-layer BiLSTM \cite{zhou-etal-2016-attention} and a nonlinear fully connected layer. After pooling, the embedding of each table and its annotation can be represented as $T_{i}, \hat{T}_{i} \in \mathbb{R}^{1 \times d} \ (i \in \{1, \ldots, N\})$, and the embeddings of each column and its annotation can be represented as $C_{i,k}, \hat{C}_{i,k} \in \mathbb{R}^{1 \times d} \ (i \in \{1, \ldots, N\}, k \in \{1, \ldots, n_{i}\})$, where $d$ denotes the size of the hidden layer.

\vspace{4pt}
\noindent\textbf{Schema Semantic Enhancement Layer} 
The naming of database schemas can sometimes be ambiguous, which may lead to a semantic discrepancy between the user's query intent and the actual data structure. As shown in Figure \ref{model}, parsing through LLM reveals that the "\textit{continent}" column in both the \textit{Continents} and \textit{Countries} tables represents "\textit{continent name}" and "\textit{continent id}" respectively. Such abbreviated column names can lead to misinterpretations, thereby affecting the performance of the schema extractor. To address this issue of semantic inconsistency, we introduce an attention gating mechanism on top of the column-enhanced module \cite{li2023resdsql} to aggregate representations of schema names along with their associated annotations.
\begin{equation}
    \mathbf{g_i^t} = \sigma(\mathbf{W}_g^t [\mathbf{W}_1^t T_{i} + \mathbf{b}_1^t; \mathbf{W}_2^t \hat{T}_{i} + \mathbf{b}_2^t] + \mathbf{b}_g^t)
\end{equation}
\begin{equation}
    T_{i}^{G} =Norm(T_{i}+(\mathbf{g_i^t} \circ T_{i}+(1-\mathbf{g_i^t})\circ \hat{T}_{i}))
\end{equation}
\begin{equation}
    \resizebox{.89\hsize}{!}{$
        \mathbf{g_{i,k}^c} = \sigma(\mathbf{W}_g^c [\mathbf{W}_1^c C_{i,k} + \mathbf{b}_1^c; \mathbf{W}_2^c \hat{C}_{i,k} + \mathbf{b}_2^c] + \mathbf{b}_g^c)
    $}
\end{equation}
\begin{equation}
    \resizebox{.89\hsize}{!}{$
        C_{i,k}^{G} =Norm(T_{i}+(\mathbf{g_{i,k}^c} \circ C_{i,k}+(1-\mathbf{g_{i,k}^c})\circ \hat{C}_{i,k}))
    $}
\end{equation}
where $ \mathbf{W}_j^t, \mathbf{W}_j^c \in \mathbb{R}^{d \times d}, \mathbf{b}_j^t, \mathbf{b}_j^c \in \mathbb{R}^{d} \, (j=1,2) $, and $ \circ $ denotes the element-wise multiplication. Additionally, $ \mathbf{b}_g^t, \mathbf{b}_g^c \in \mathbb{R}^{d} $, $ \mathbf{W}_g^t, \mathbf{W}_g^c \in \mathbb{R}^{2d \times d} $, and $ \sigma(\cdot) $ represents the sigmoid function, while $ Norm(\cdot) $ is the row-wise $ L_2 $ normalization function. The gating vectors $ \mathbf{g_i^t} $ and $ \mathbf{g_{i,k}^c} $ represent the gating vectors for tables and columns, respectively. By using the probabilities from these gating vectors to perform a weighted average and normalization of the embeddings of schemas and their annotations, we obtain the enhanced table embeddings $ T_{i}^{G} \in \mathbb{R}^{1 \times d} $ and column embeddings $ C_{i,k}^{G} \in \mathbb{R}^{1 \times d} $.

After processing through the semantic enhancement layer, we obtain the classification probabilities of all schemas for the current turn. Next, based on a predefined threshold $s$, we extract the schemas that are most relevant to the question. This threshold must be set reasonably to avoid losing important schemas due to it being too high. We sort the selected schemas in descending order of their probability values and combine them with the serialized foreign key information, ultimately generating the serialized database schema representation required by the SQL generation model. 

\vspace{4pt}
\noindent\textbf{ALL-Column Intent Detection} 
In question interactions, user intent is not always directly expressed as seeking information from specific columns but may implicitly involve retrieving information from all columns within a specified table. To effectively capture the implicit "all-columns intent" within user questions, we designed the schema extractor to specifically recognize the wildcard "*" in SQL as a special column identifier. For example, in the question "\textit{Show ids and names of all continents!}", the user explicitly requests information from specific columns ("\textit{ids}" and "\textit{names}") in the "\textit{continents}" table. However, when dealing with broader questions such as "\textit{Which countries do they each have}", the user's intent implicitly demands retrieving all relevant data. In response to this, the schema extractor generates classification probabilities for "*" across all tables and determines whether and how to insert "*" into the input sequence based on these probabilities. This approach guides the SQL generator to more accurately understand and respond to the user's question intent.

\subsection{Schema-aware Context Extractor}
\label{sec:History SQL Selection}
As the number of dialogue turns increases, the relationships between each turn become more intricate, with current questions often bearing inter-turn connections to historical questions. To address this, we have developed a Schema-aware Context Extractor that can identify the most relevant past SQL related to the current question and use the query as references for generating the current SQL. Since semantic associations between questions and the database schema have already been modeled during the schema extraction phase, in the subsequent context extraction process, we can directly utilize the pre-stored schema item encoding information in the history schema store, thereby avoiding redundant model training.

In the process of selecting historical question-SQL pairs, we mainly consider the following two factors. Firstly, the semantic relevance between the current question $\mathcal{Q}_{m}$ and historical questions $\mathcal{Q}_{h} \ (h \in 1,...,m-1)$. Higher semantic similarity typically indicates similar query intentions. For example, if $\mathcal{Q}_{m}$ is \textit{"How many dog pets are raised by female students?"}, and the historical question $\mathcal{Q}_{h}$ is \textit{"How many of those have dogs?"}, there is a clear semantic association between these questions. We employ the SentenceBERT \cite{reimers-gurevych-2019-sentence} to quantify this similarity, defined as:
\begin{equation}
    \mathcal{S}^{sim}_{h} = SentenceBERT(\mathcal{Q}_{h}, \mathcal{Q}_{m})
\end{equation}
where $ h\in[1,..,m-1] $. When $ \mathcal{Q}_{h} $ and $ \mathcal{Q}_{m} $ are semantically similar, their corresponding SQL queries likely share a similar structure.

Secondly, relying solely on semantic similarity between questions might lead to misjudgment. For instance, the questions \textit{"Who are the female students?"} and \textit{"Of those, who has a pet?"} are sequential inquiries but appear quite different semantically. Therefore, we also incorporate the schema item probabilities provided by the History Schema Store to measure the overlap of entities involved in the two questions. Let the normalized schema item extraction probability vectors obtained from the $ m^{th} $ and $ h^{th} $ turns be denoted as $\hat{\mathcal{Y}}^{m}$ and $\hat{\mathcal{Y}}^{h}$, respectively. The normalized Jensen-Shannon divergence can then be used to assess the difference between these entities:
\begin{equation}
    \resizebox{.89\hsize}{!}{
        $\mathcal{P}^{sim}_{h} = \frac{1}{2\ln{2}} (D_{KL}(\hat{\mathcal{Y}}^{m} || \bar{\mathcal{Y}}) + D_{KL}(\hat{\mathcal{Y}}^{h} || \bar{\mathcal{Y}}))$
    }
\end{equation}
where $\bar{\mathcal{Y}} = \frac{1}{2} (\hat{\mathcal{Y}}^{m} + \hat{\mathcal{Y}}^{h})$ is the average distribution of $\hat{\mathcal{Y}}^{m}$ and $\hat{\mathcal{Y}}^{h}$, and $ D_{KL} $ represents the Kullback-Leibler divergence, which can be expressed as:
\begin{equation}
    \begin{aligned}
        &D_{KL}(\hat{\mathcal{Y}}^{m} || \hat{\mathcal{Y}}^{h}) = \\
        &\sum_{i=1}^{N} \left( \hat{y}_{i}^{h} \log \frac{\hat{y}_{i}^{h}}{\hat{y}_{i}^{m}} + \sum_{k=1}^{n_{i}} \hat{y}_{i,k}^{h} \log \frac{\hat{y}_{i,k}^{h}}{\hat{y}_{i,k}^{m}} \right)
    \end{aligned}
\end{equation}
where $\hat{y}_{i}^{h}$ denotes the normalized predicted probability of the $ i^{th} $ table in the $ h^{th} $ turn, while $\hat{y}_{i,k}^{h}$ represents the normalized predicted probability of the $ k^{th} $ column within that table.

$\mathcal{S}^{sim}_{h}$ and $\mathcal{P}^{sim}_{h}$ are two key scoring metrics used for retrieving and filtering historical information, both of which are constrained within the interval [0,1]. Specifically, $\mathcal{S}^{sim}_{h}$ represents a maximization score metric, whereas $P^{sim}_{h}$ is a minimization score metric. To unify the direction of the metrics, we adjust the form of $\mathcal{P}^{sim}_{h}$ to $1 - \mathcal{P}^{sim}_{h}$, thereby allowing both metrics to be optimized towards maximization. Based on these definitions, we can calculate the comprehensive relevance score $\mathcal{R}_{h}$ between $\mathcal{Q}_{h}$ and $\mathcal{Q}_{m}$ as follows:
\begin{equation}
    \mathcal{R}_{h} = \mathcal{S}^{sim}_{h} + 1 - \mathcal{P}^{sim}_{h}
\end{equation}

By selecting historical SQL with the highest $\mathcal{R}_{h}$ values, we can utilize appropriate historical SQL as reference inputs during the supervised fine-tuning of an LLM. This facilitates generating accurate SQL outputs in response to $Q_{m}$, effectively reducing the discrepancy between inputs and outputs.

\subsection{SQL Generation Fine-tuning}
\label{sec:SQL Generation Fine-tuning}
By screening the schemas in multiple turns of data sets and filtering out irrelevant historical information, we obtain a streamlined input sequence that is highly relevant to the target SQL. Based on this, we transform the multi-turn dataset into a single-turn Text-to-SQL corpus. Each entry's input sequence consists of the question $\mathcal{Q}_{\le m}$, the extracted sequence of schemas $E(\mathcal{S})$, and $SQL_{base}$, while the actual SQL serves as the desired output sequence $s_{m}$. 
In practice, when converting multi-turn questions into single-turn questions, we found that directly rewriting the questions or other forms of filtering could lead to severe error propagation. Therefore, we define the problem $\mathcal{Q}_{\le m} = \mathcal{Q}_{1} \& ... \& \mathcal{Q}_{m}$, using symbols to concatenate the sequences to ensure the semantic integrity of the multi-turn question series. For the first question $\mathcal{Q}_{1}$, due to the lack of a previous SQL as a reference, we set $SQL_{base}$ as an empty sequence.
Therefore, the minimization loss function of a set of interaction samples can be expressed as:
\begin{equation}
    \min_{\varepsilon ,M^{*}} \sum_{m}^{} \mathcal{L}(M^{*}\varepsilon  (\mathcal{Q}_{\le m},E(\mathcal{S}),SQL_{base},s_{m}))
\end{equation}
where $\varepsilon(\cdot)$ defines a sequence format and details can be found in Appendix \ref{sec:Format for fine-tuning SQL generation}.
Due to the Track-SQL framework enabling dynamic schema linking and effective filtering of historical information, the LLM can focus more on capturing the essential connections between key information and SQL queries during training, thereby reducing the interference caused by redundant information in the SQL generation task.

\section{Experiments}

\subsection{Experimental Setup}

\noindent\textbf{Datasets}
We validated the effectiveness of the proposed method on the SParC \cite{yu-etal-2019-sparc} and CoSQL \cite{yu-etal-2019-cosql} benchmark datasets. The SParC dataset contains 4,298 multi-turn dialogue sequences, covering over 12,000 individual questions and their corresponding SQL queries. The CoSQL dataset includes over 10,000 annotated SQL queries, with each dialogue sequence designed to simulate real-world scenarios where ordinary users explore databases and interact with them. In these scenarios, non-expert users ask questions in natural language, while experts use SQL to retrieve answers. 

\vspace{4pt}
\noindent\textbf{Evaluation Metrics}
To evaluate the performance of our method in the text-to-SQL task, we adopted two official metrics: Question Match (QM) and Interaction Match (IM), along with three widely recognized sub-metrics: Exact Match Accuracy (EM), Execution Accuracy (EX), and Test Suite Accuracy (TS). QM measures whether the predicted SQL is accurate for a single question, while IM evaluates whether all predicted SQL queries meet the QM standard across multiple turns of a conversation. Specifically, EM measures the structural accuracy of the predicted SQL; EX focuses on whether the execution result of the predicted SQL is correct; TS not only examines the accuracy of query execution but also requires correct results for each query executed over multiple database instances and schemas. 

To provide a finer-grained assessment of the extraction accuracy of database schemas, we introduced a set of strict evaluation metrics: Table Redundancy Score ($TRS@s$) and Column Redundancy Score ($CRS@s$), where $s$ represents the extraction threshold. A schema item is considered for extraction when its classification probability calculated by the schema extractor is greater than $s$. These metrics are formally defined as follows:
\begin{equation}
    V^{t}=\{{i}' \ |\  {y}_{{i}'}=1  \},\ \hat{V}^{t}=\{{i}' \ | \ \hat{y}_{{i}'}\ge s\}
\end{equation}
where $V^{t}$ denotes the set of indices of tables included in the ground truth SQL, and $\hat{V}^{t}$ denotes the set of indices of tables extracted by the schema extractor. The score $score_{j}$ is calculated based on the ratio of redundant elements to total elements in $\hat{V}^{t}$, defined as:
\begin{equation}
    \left.score_{j}=\left\{\begin{matrix}0& if \; V^{t}=\hat{V}^{t}
     \\\frac{|\hat{{V}}^{t} - V^{t}|}{|\hat{V}^{t}|}&if \; V^{t}\subset \hat{V}^{t} \\1 & if \; V^{t}\not\subset \hat{V}^{t} \end{matrix}\right.\right.
\end{equation}
\begin{equation}
    TRS@s=\frac{1}{D} \sum_{j=1}^{D} score_{j}
\end{equation}
where $|\cdot|$ denotes the number of elements in the set. $TRS@s$ is obtained by averaging $score_{j}$, where $D$ represents the total number of samples. For $CRS@s$, we use a similar approach, combining the set of indices of columns included in the ground truth SQL with the set of indices of columns extracted by the schema extractor to compute the score $score_{i,j}$. Here, $score_{i,j}$ represents the redundancy of the set of extracted columns in the $i^{th}$ table referred to the $j^{th}$ sample. By summing these scores, we can obtain the $CRS@s$ for column name extraction, where $N_{j}$ denotes the number of tables contained in the $j^{th}$ sample's database.
\begin{equation}
    CRS@s=\frac{1}{\sum_{j=1}^{D}N_{j}  } \sum_{j=1}^{D} \sum_{i=1}^{N_{j}} score_{j}^{i}
\end{equation}

\vspace{4pt}
\noindent\textbf{Implementation Details}
To generate descriptive annotations for the database schemas, we utilized the GPT-3.5-turbo model\footnote{\url{https://openai.com/}} and provided detailed input prompts in Appendix \ref{sec:annotation_prompt}. During the training of the schema extraction model, we employed the AdamW optimizer \cite{loshchilov2019decoupledweightdecayregularization}.
For the training phase of the SQL generator, we used the transformers library for LoRA fine-tuning \cite{hu2021loralowrankadaptationlarge}. The specific configurations were as follows: LoRA rank was set to 32, LoRA alpha was set to 64, and LoRA dropout was set to 0.1. The batch size for training was set to 6.

In the experimental configuration of the Track-SQL framework, systematic settings were implemented for three key aspects: schema linking stability, model input constraints, and context management. The schema item sequence perturbation mechanism applies random reshuffling and irrelevant item insertion to schema items with probability 0.15 during training, while employing dynamic filtering with an extraction threshold \(s=0.1\) during inference. To accommodate the 512-token input limit of the Roberta base model, a column-level schema segmentation strategy divides large databases into column-unit-based sub-schema sets that conform to length constraints. For context expansion management, we designed:
\begin{itemize}
    \item A fixed-length sliding window (\(L_{w}=5\))
    \item A queue-based update rule: When context unit \(c_t\) arrives, append it if window capacity permits; otherwise remove the earliest unit \(c_{t-L_{w}}\) before inserting \(c_t\)
\end{itemize}

All experiments were conducted on a high-performance server configured with an NVIDIA A800 (80GB) GPU, a Hygon C86 7390 32-core processor, 2TB of memory, and the Ubuntu 22.04.3 LTS operating system.

\subsection{Main Results}
\begin{table*}[ht]
    \centering
    \large
    \renewcommand\arraystretch{1.2}
    \resizebox{\textwidth}{!}{
        \begin{tabular}{c|ccc|ccc|ccc|ccc}
            \toprule [0.5mm] 
            \multirow{3}{*}{Model} & \multicolumn{6}{c|}{SparC} & \multicolumn{6}{c}{CoSQL} \\
            \cline{2-13} 
             & \multicolumn{3}{c|}{QM} & \multicolumn{3}{c|}{IM} & \multicolumn{3}{c|}{QM} & \multicolumn{3}{c}{IM} \\
             \cline{2-13}
             & EM & EX & TS & EM & EX & TS & EM & EX & TS & EM & EX & TS \\
            \midrule
            \multicolumn{13}{c}{\textbf{In-Context Learning Approach}} \\
            \midrule
            ACT-SQL \cite{zhang-etal-2023-act} & 51.0 & 63.8 &56.9 & 24.4 & 38.9 &29.6 & 46.0 & 63.7 & 55.2 & 13.3 & 30.7 & 21.5 \\
            CoE-SQL \cite{zhang-etal-2024-coe} & 56.0 & 70.3 & 63.3 & 36.5 & 50.5 & 41.9 & 52.4 & 69.6 & 60.6 & 23.9 & 39.6 & 30.4 \\
            \midrule
            \multicolumn{13}{c}{\textbf{Fine-tuned Model}} \\
            \midrule
            HIE-SQL + GraPPA \cite{zheng-etal-2022-hie} & 64.7 & - & - & 45.0 & - & - & 56.4 & - & - & 28.7 & - & - \\
            RASAT + PICARD \cite{qi-etal-2022-rasat} & 67.7 & 73.3 & - & 49.1 & 54.0 & - & \textbf{58.8} & 67.0 & - & 27.0 & 39.6 & - \\
            QDA-SQL \cite{sun2024qdasqlquestionsenhanceddialogue} & 61.3 & - &- & 44.1 & - &- & 57.3 & - & - & \textbf{30.0} & - & - \\
            \midrule
            \multicolumn{13}{c}{\textbf{Ours}} \\
            \midrule
            SFT Codellama 7B & 59.18 & 67.99 & 61.51 & 36.96 & 46.68 & 39.33 & 50.54 & 60.17 & 54.51 & 17.74 & 28.66 & 23.54 \\
            + Track-SQL & 61.26 & 70.07 & 63.75 & 43.36 & 52.13 & 45.73 & 53.82 & 66.73 & 59.18 & 24.57 & 37.88 & 29.01 \\
            \cdashline{1-13}
            SFT Deepseek 7B & 64.33 & 71.40 &65.08 & 43.36 & 50.71 & 43.36 & 54.71 & 66.03 & 58.88 & 23.20 & 34.12 & 26.96\\
            + Track-SQL & 65.17 & \textbf{75.39} & \textbf{69.16} & 46.44 & \textbf{57.81} & \textbf{50.71} & 58.19 & 70.60 & 62.26 & 28.67 & 43.67 & 32.76 \\
            \cdashline{1-13}
            SFT Mistral 7B & 64.17 & 70.82 & 65.58 & 43.60 & 52.13 & 45.49 & 56.20 & 64.94 & 59.68 & 24.57 & 34.81 & 29.01 \\
            + Track-SQL & \textbf{65.41} & 73.23 & 67.83 & \textbf{46.91} & 54.73 & 48.57 & 57.69 & \textbf{71.10} & \textbf{65.54} & 27.30 & \textbf{45.05} & \textbf{36.17} \\
            \bottomrule [0.5mm]
        \end{tabular}
    }
    \caption{Performance of Track-SQL and previous works on the SparC and CoSQL dev set.}
    \label{tab:result}
\end{table*}

During the experimental process involving multiple turns of SQL generation, we selected three representative models of 7B scale: CodeLlama\footnote{\url{https://huggingface.co/codellama/CodeLlama-7b-Instruct-hf}}, 
DeepSeek\footnote{\url{https://huggingface.co/deepseek-ai/deepseek-coder-6.7b-instruct}}, and Mistral\footnote{\url{https://huggingface.co/mistralai/Mistral-7B-Instruct-v0.3}}. We directly utilized the multi-turn questions and their database schemas from the SparC and CoSQL datasets as inputs to train and infer using these 7B-scale models, thereby obtaining baseline results. Subsequently, we applied the Track-SQL method to optimize the inputs and retrained and inferred with the models. By comparing the experimental results under the Track-SQL framework with the baseline results, we obtained the detailed comparative data presented in Table \ref{tab:result}.

In terms of single-turn and multi-turn evaluation metrics, the 7B-scale models under the Track-SQL framework significantly outperformed the baseline models, validating the effectiveness of our input optimization method. Specifically, under the DeepSeek 7B base model, Track-SQL improved the single-turn EX metric by 3.99$\%$ and the TS metric by 4.08$\%$ on the SparC development set; it also improved the multi-turn EX metric by 7.1$\%$ and the TS metric by 7.35$\%$. Similar improvements were confirmed within the CoSQL dataset. These results indicate that the Track-SQL method is highly effective in handling multi-turn interactive question answering tasks.

We also categorized other multi-turn Text-to-SQL research efforts according to different inference methodologies into two classes: In-context learning methods and fine-tuning methods. Compared to these methods, Track-SQL achieved the best performance on the SparC and CoSQL development sets, surpassing previous In-context learning and fine-tuning techniques under both single-turn and multi-turn evaluation metrics. Specifically, compared to In-context learning methods, such as ACT-SQL and CoE-SQL which rely on GPT 3.5, the fine-tuning experimental results under the Track-SQL framework with 7B-scale models were superior. Unlike In-context learning methods which can be influenced by historical questions, fine-tuning methods are better at focusing on the core of the current turn's question. Moreover, even though RASAT+PICARD combines beam search strategies for SQL correction, 
Track-SQL achieved notably higher multi-turn inference accuracies on the SparC and CoSQL development sets without employing any SQL post-processing techniques.

\subsection{Ablation Studies}
\begin{table}
    \centering
    \large
    \renewcommand\arraystretch{1.3}
    \resizebox{\linewidth}{!}{
        \begin{tabular}{c|cc|cc}
            \toprule 
             & \multicolumn{2}{c|}{\textbf{QM}} & \multicolumn{2}{c}{\textbf{IM}} \\
             & \textbf{EX} & \textbf{TS} & \textbf{EX} & \textbf{TS} \\
            \midrule
            \textbf{Track-SQL} & \textbf{75.39} & \textbf{69.16} & \textbf{57.81} & \textbf{50.71} \\
            w/o SESE & \cellcolor[HTML]{ebebff}68.57\small{($\downarrow{6.82}$)} & \cellcolor[HTML]{ebebff}62.42\small{($\downarrow{6.74}$)} & 51.42\small{($\downarrow{6.39}$)} & 45.26\small{($\downarrow{5.45}$)} \\
            w/o ACID & 74.56\small{($\downarrow{0.83}$)} & 67.91\small{($\downarrow{1.25}$)} & 56.39\small{($\downarrow{1.42}$)} & 48.57\small{($\downarrow{2.14}$)} \\
            w/o SACE & 72.73\small{($\downarrow{2.66}$)} & 67.16\small{($\downarrow{2}$)} & 51.89\small{($\downarrow{5.92}$)} & 44.78\small{($\downarrow{5.93}$)} \\
            w/o SACE \& SESE & 71.40\small{($\downarrow{3.99}$)} & 65.08\small{($\downarrow{4.08}$)} & \cellcolor[HTML]{ebebff}50.71\small{($\downarrow{7.10}$)} & \cellcolor[HTML]{ebebff}43.36\small{($\downarrow{7.35}$)} \\
            \bottomrule
        \end{tabular}
    }
    \caption{Ablation result on SparC dev set.}
    \label{tab:ablation_result_on_sparc}
\end{table}

\begin{table}
    \centering
    \large
    \renewcommand\arraystretch{1.3}
    \resizebox{\linewidth}{!}{
        \begin{tabular}{c|cc|cc}
            \toprule 
             & \multicolumn{2}{c|}{\textbf{QM}} & \multicolumn{2}{c}{\textbf{IM}} \\
             & \textbf{EX} & \textbf{TS} & \textbf{EX} & \textbf{TS} \\
            \midrule
            \textbf{Track-SQL} & \textbf{70.60} & \textbf{62.26} & \textbf{43.67} & \textbf{32.76} \\
            w/o SESE & \cellcolor[HTML]{ebebff}64.94\small{($\downarrow{5.66}$)} & 58.98\small{($\downarrow{3.28}$)} & 36.86\small{($\downarrow{6.81}$)} & 31.74\small{($\downarrow{1.02}$)} \\
            w/o ACID & 68.22\small{($\downarrow{2.38}$)} & 61.37\small{($\downarrow{0.89}$)} & 39.24\small{($\downarrow{4.43}$)} & 32.42\small{($\downarrow{0.34}$)} \\
            w/o SACE & 68.81\small{($\downarrow{1.79}$)} & 61.46\small{($\downarrow{0.8}$)} & 37.88\small{($\downarrow{5.79}$)} & 29.01\small{($\downarrow{3.75}$)} \\
            w/o SACE \& SESE & 66.03\small{($\downarrow{4.57}$)} & \cellcolor[HTML]{ebebff}58.88\small{($\downarrow{3.38}$)} & \cellcolor[HTML]{ebebff}34.12\small{($\downarrow{9.55}$)} & \cellcolor[HTML]{ebebff}26.96\small{($\downarrow{5.8}$)} \\
            \bottomrule
        \end{tabular}
    }
    \caption{Ablation result on CoSQL dev set.}
    \label{tab:ablation_result_on_cosql}
\end{table}

To further validate the effectiveness of the proposed method, we conducted detailed ablation studies on the development sets of Sparc and CoSQL to evaluate the importance of each module. Tables \ref{tab:ablation_result_on_sparc} and \ref{tab:ablation_result_on_cosql} present the experimental results based on the Deepseek 7B model. Additionally, in Appendix \ref{sec:ablation_by_differentLLMs}, we provide the results of extra ablation experiments conducted using models of different 7B parameter scales, thereby confirming that the performance improvements attributed to each module are not due to the randomness associated with any particular 7B model. To further analyze how each module enhances model performance, we categorized the datasets according to the complexity of the SQL statements and the number of dialogue interactions, and conducted targeted ablation experiments accordingly. The detailed results are shown in the appendix \ref{sec:ablation_by_difficulty} and \ref{sec:ablation_by_turn}. By labeling the table columns included in the ground truth SQLs of the multi-turn conversation datasets, we performed ablation experiments on the schema extractor. The results are shown in Table \ref{tab:ablation_result_schema_extractor}, confirming the effectiveness of this module in reducing redundancy during the schema item extraction process.

\vspace{4pt}
\noindent\textbf{Effect of Semantic-enhanced Schema Extractor} We employ a two-phase evaluation metric to assess the effectiveness of SESE. As shown in Table \ref{tab:ablation_result_schema_extractor}, the introduction of the semantic enhancement module in the extractor significantly reduces redundancy rates in table and column extraction. This improvement is particularly pronounced on the CoSQL dataset, which contains a higher proportion of multi-turn ambiguous dialogue samples compared to the SparC dataset, posing more stringent challenges for schema understanding mechanisms. Through LLM-based contextual semantic expansion and annotation embedding mechanisms, this module effectively mitigates schema comprehension biases in complex dialogue turns. Specifically, on CoSQL dev set, $TRS@0.5$ and $CRS@0.5$ decrease by approximately 0.88\% and 4.03\% respectively, demonstrating substantial progress in schema item localization accuracy. Additional ablation study results under alternative classification metrics are provided in Appendix \ref{sec:comprehensive_eval_schema_item_extractor}, further validating the critical role of the semantic enhancement module.

During SQL generation, when removing SESE module, QM-TS and IM-TS metrics decline by 6.74\% and 5.45\% respectively. This not only confirms the strong task coupling between schema extraction and SQL generation, but also verifies our core hypothesis: optimizing the precision-redundancy balance in schema extraction can effectively enhance multi-turn SQL generation performance, thereby establishing the theoretical feasibility of our proposed approach.

\begin{table}[]
    \centering
    \large
    \renewcommand\arraystretch{1.3}
    \resizebox{\linewidth}{!} {
        \begin{tabular}{c|cc|cc}
            \toprule 
             & \multicolumn{2}{c|}{\textbf{SparC}} & \multicolumn{2}{c}{\textbf{CoSQL}} \\
             & $TRS@0.5$ & $CRS@0.5$ & $TRS@0.5$ & $CRS@0.5$ \\
             \midrule
             \textbf{Track-SQL} & \textbf{8.50} & 21.46 & \textbf{12.74} & \textbf{25.90} \\
             w/o com-enh & 9.57 & \textbf{21.05} & 13.62 & 29.93 \\
             w/o com\&col-enh & 9.44 & 28.18 & 13.38 & 27.42 \\
             \bottomrule
        \end{tabular}
    }
    \caption{Ablation results of SESE. com-enh represents comment-enhanced layer, and col-enh represents column-enhanced layer used in RESDSQL.}
    \label{tab:ablation_result_schema_extractor}
\end{table}

\vspace{4pt}
\noindent\textbf{Effect of All-Column Intent Detection} When the All-Column Intent Detection (ACID) strategy is eliminated, all metrics for Tables \ref{tab:ablation_result_on_sparc} and \ref{tab:ablation_result_on_cosql} exhibit slight performance degradation. Combined with the case study in Appendix \ref{sec:case study}, this verifies that the ACID strategy possesses limited yet discernible capability in enhancing the model's recognition of user all-column intentions. In the inference experiments on the COSQL dev set, removing this module resulted in decreases of 0.89\% and 0.34\% in QM-TS and IM-TS respectively, while more significant drops of 2.38\% and 4.43\% were observed in QM-EX and IM-EX. This phenomenon suggests that the EX evaluation metrics contain a higher proportion of false positive samples during their calculation process. The optimization effect of this strategy on the overall performance of the Track-SQL framework demonstrates metric sensitivity characteristics. Final experimental data confirm that this module yields only limited performance gains in real-world application scenarios.

\vspace{4pt}
\noindent\textbf{Effect of Schema-aware Context Extractor} The strategy of employing historical SQL as prompt generation exhibits inherent limitations: when structural errors exist in historically generated statements, erroneous prompts may trigger cascading error propagation. To mitigate this, SACE incorporates syntactic error detection in historical SQL generation to reduce error transmission. Ablation experiments demonstrate that removing SACE module resulted in performance degradation of 5.92\% and 5.93\% on IM-EX and IM-TS metrics respectively in the SparC validation set, exhibiting more pronounced degradation compared to QM metrics. This phenomenon indicates that under relatively high model accuracy conditions, the SACE module significantly enhances the completeness of multi-turn SQL generation. Notably, the marginal effects revealed in Appendix \ref{sec:ablation_by_difficulty} demonstrate that when query difficulty escalates to EXTRA levels, SACE inadvertently induces error propagation. This revelation exposes the module's limitations, thereby suggesting that achieving more robust multi-turn SQL generation systems requires the implementation of systematic validation strategies.

\subsection{Execution Time Analysis}
To evaluate the practical performance of the Track-SQL framework, we conducted latency analysis and training efficiency validation on its core components. As shown in Table \ref{tab:time_study}, experimental results indicate that under typical multi-turn database query scenarios, the system achieves an end-to-end response time of 1.35 seconds (0.20 seconds for SESE and 1.15 seconds for SQL Generator), satisfying the latency requirements for real-time interactive systems.Regarding training efficiency, under standard experimental configurations on the SparC dataset, the schema extractor achieves optimal performance after 30.9 hours of training, while the SQL generator converges to its optimal state in approximately 1.5 hours. This demonstrates the framework's significant efficiency advantages in computational resource utilization.

\begin{table}
    \centering
    \large
    \renewcommand\arraystretch{1.3}
    \resizebox{\linewidth}{!}{
        \begin{tabular}{c|cc|cc}
            \toprule 
             & \multicolumn{2}{c|}{\textbf{SESE}} & \multicolumn{2}{c}{\textbf{SQL Generator}} \\
             & \textbf{Train} & \textbf{Inference} & \textbf{Train} & \textbf{Inference} \\
            \midrule
            SparC & 30.9±2.8(h) & 0.20(s) 
                  & 1.5±0.2(h) & 1.15(s) \\
            CoSQL & 27.5±2.4(h) & 0.21(s) 
                  & 1.6±0.3(h) & 1.15(s) \\
            \bottomrule
        \end{tabular}
    }
    \caption{Time performance of the Track-SQL framework, \textbf{Train} employs the number of hours required to achieve the optimal model as the performance metric, while \textbf{Inference} bases its evaluation on the time consumed to perform inference on a single batch.}
    \label{tab:time_study}
\end{table}

\section{Related Work}
\noindent\textbf{Schema Linking} 
Schema linking effectively reduces errors caused by schema misinterpretation, thereby enhancing the performance of text-to-SQL conversion. Currently, various methods focus on improving schema linking. For instance, DIN-SQL \cite{li2024can} utilizes GPT to identify relevant database elements, while DTS-SQL \cite{pourreza2024dtssqldecomposedtexttosqlsmall} employs a specialized model for efficient schema extraction. RESDSQL \cite{li2023resdsql} and CodeS \cite{li2024codesbuildingopensourcelanguage} adopt strategies that assess the relevance between schemas and queries for sorting and filtering. Other methods, such as C3 \cite{dong2023c3zeroshottexttosqlchatgpt}, CHESS \cite{talaei2024chesscontextualharnessingefficient}, and MCS-SQL \cite{lee2024mcssqlleveragingmultipleprompts}, adopt a step-by-step linking approach, first filtering out relevant tables and then selecting matching columns. Existing methods perform well in single-turn Text-to-SQL tasks, but they are inadequate in handling multi-turn dialogues and schema name ambiguities. This paper focuses on dynamic schema linking in multi-turn Text-to-SQL, with a particular emphasis on resolving schema name ambiguities. Our proposed method reduces redundant links and improves system performance.

\vspace{4pt}
\noindent\textbf{Multi-turn Text-to-SQL }
Multi-turn text-to-SQL tasks more closely resemble real-world applications, allowing users to progressively refine their questions and adjust their requirements through dialogue. Research in this field includes ISTSQL \cite{wang2021tracking}, which improves accuracy by tracking the states of database schemas and SQL keywords; MIGA \cite{fu2023miga} utilizes multi-task learning to integrate information on reference relationships and schema links; CoE-SQL \cite{zhang-etal-2024-coe} tracks user intent by serializing changes in SQL queries. Inspired by copying mechanisms, methods like EditSQL \cite{zhang-etal-2019-editing}, refer to previous-turn SQL information. R2SQL \cite{hui2021dynamic} introduces a memory decay mechanism to simulate changes in the database schema. TP-Link \cite{liu-etal-2024-tp} models word-level coreference to resolve complex coreference and ellipsis issues. These studies aim to support the Track-SQL framework by capturing dependencies between multi-turn text-to-SQL interactions. The framework improves the handling of multi-turn co-reference issues and enhances system performance and adaptability by integrating dynamic schema element awareness with question semantic information.

\section{Conclusion}
\label{sec:bibtex}
In this paper, we proposed the Track-SQL framework to address the challenges of multi-turn Text-to-SQL tasks, focusing on dynamic schema linking and effective utilization of historical context. Track-SQL integrates a \emph{Semantic-enhanced Schema Extractor} and a \emph{Schema-aware Context Extractor} to precisely capture schema and contextual changes, improving the system’s ability to adapt to evolving user interactions. The core idea lies in establishing the association model between user intent, schemas, and contextual information in advance, thereby reducing the difficulty of recognizing changing intents during the SQL generation process. Experimental results indicate that the Track-SQL possesses significant advantages and effectiveness.

\section*{Limitations}
In the schema extraction phase, using RoBERTa as the base model limits the maximum window length to 512, which increases the training time for handling large volumes of text data; even with an A800 (80G) GPU-equipped machine, training on the SparC dataset takes two days. We attempted to replace RoBERTa with a decoder-only model that supports a larger window size for classifier training, but the results were unsatisfactory. While the Track-SOL framework has advanced dynamic schema linking and context information extraction, its efficacy in extremely complex multi-turn dialogues and highly dynamic database schemas remains to be validated. Future efforts will focus on enhancing the framework's robustness to ensure high-performance and stability across diverse application scenarios.

\section*{Acknowledgments}
This research was financially supported by the Natural Science Foundation of China (62406078), the Open Research Fund from Guangdong Laboratory of Artificial Intelligence and Digital Economy (SZ) under Grant No. GML-KF-24-23, the National Science and Technology Major Project (2021ZD0111501), the National Science Fund for Excellent Young Scholars (62122022), the Natural Science Foundation of China (U24A20233, 62476163, 62206064, 62206061), the major key project of PCL (PCL2021A12), the Guangdong Basic and Applied Basic Research Foundation (2023B1515120020), and the Collaborative Education Project of the Ministry of Education (202407). This research was enabled by the computational resources and support of the High Performance Computing Platform at the School of Computer Science, Guangdong University of Technology.

\bibliography{anthology,custom}

\appendix

\section{Appendix A}
\label{sec:appendix a}

This section provides additional experimental data regarding the proposed Track-SQL framework, detailing the ablation study results categorized by different large language models, SQL complexity, and number of dialogue turns. These results further confirm the effectiveness and performance of the designed schema extractor and SQL generator. Specifically, in section \ref{sec:ablation_by_difficulty}, we present the ablation study results of Track-SQL across different levels of SQL difficulty. Section \ref{sec:ablation_by_turn} evaluates the framework's performance under varying numbers of dialogue turns. In Section \ref{sec:ablation_by_differentLLMs}, we selected three mainstream open-source LLM models and conducted ablation experiments on the SparC and CoSQL datasets. Section \ref{sec:comprehensive_eval_schema_item_extractor} reports the performance of the \emph{Schema Extractor} across multiple classification metrics. Section \ref{sec:case study}, we conduct an in-depth analysis of specific ablation study cases. Finally, Section \ref{sec:resource_overhead_experiment} reports the performance of Track-SQL in terms of time and cost metrics. 


\subsection{Further Ablation Studies}
\label{sec:further_ablation_studies}

\subsubsection{Ablation by difficulty}
\label{sec:ablation_by_difficulty}
In the Spider \cite{yu2018spider} dataset, SQL queries are categorized into four levels: Easy, Medium, Hard, and Extra. The difficulty level is determined by the number of SQL components, selections, and conditions. Queries containing more SQL keywords (such as \textit{GROUP BY}, \textit{ORDER BY}, \textit{INTERSECT}, \textit{nested subqueries}, \textit{multi-column selections}, and \textit{aggregation operations}) are considered to be of higher difficulty. Specifically, if a query includes more than two \textit{SELECT} columns, more than two \textit{WHERE} conditions, uses \textit{GROUP BY} on two or more columns, or contains \textit{EXCEPT} keywords or nested queries, it is classified as \textbf{Hard}. Queries that further increase complexity beyond this are classified as \textbf{Extra}. The SparC and CoSQL dev sets also follow this standard for categorizing sample difficulty.

We present the ablation results of Track-SQL on SparC and CoSQL dev sets in Tables \ref{tab:ablation_by_difficulty_sparc} and \ref{tab:ablation_by_difficulty_cosql}, categorized by difficulty. The results indicate:
\begin{enumerate}
    \item As the complexity of SQL reasoning increases, the SACE method—using historical base SQL to infer the SQL for the current problem—shows limitations, particularly when handling high-difficulty samples in the CoSQL dev set, where the base SQL might mislead large language models (LLMs). However, for medium-difficulty SQL, this method positively impacts SQL generation.
    \item Overall, the ACID module provides a minor improvement in SQL generation performance across all difficulty levels. Combined with case analysis (\ref{sec:case study}), this demonstrates that the ACID module successfully enhances the recognition of all columns involved in user intent, regardless of sample difficulty.
    \item The SESE module significantly improves SQL generation performance across different difficulty levels in both datasets. This is attributed to the critical role of schema linking in various types of SQL queries. Accurate schema information is fundamental to achieving high-precision SQL generation.
\end{enumerate}

\begin{table*}[]
    \centering
    \renewcommand\arraystretch{1.4}
    \resizebox{\textwidth}{!}{
        \begin{tabular}{c|ccc|ccc|ccc|ccc}
             \toprule[0.5mm]
             & \multicolumn{3}{c|}{Easy (483)} & \multicolumn{3}{c|}{Medium (441)} & \multicolumn{3}{c|}{Hard (145)} & \multicolumn{3}{c}{Extra (134)}\\
             & EM & EX & TS & EM & EX & TS & EM & EX & TS & EM & EX & TS \\
             \midrule
             Track-SQL & \textbf{81.8} & \textbf{84.9} & \textbf{83.0} & \textbf{64.6} & \textbf{72.8} & \textbf{66.4} & \textbf{44.8} & \textbf{64.8} & \textbf{53.8} & \textbf{35.1} & \textbf{54.5} & \textbf{38.8} \\
             w/o SESE & \cellcolor[HTML]{ebebff}77.6 & \cellcolor[HTML]{ebebff}78.7 & \cellcolor[HTML]{ebebff}77.6 & \cellcolor[HTML]{ccccff}56.5 & \cellcolor[HTML]{ebebff}68.0 & \cellcolor[HTML]{ccccff}59.4 & 42.8 & \cellcolor[HTML]{ebebff}59.3 & \cellcolor[HTML]{ccccff}46.2 & \cellcolor[HTML]{ebebff}30.6 & \cellcolor[HTML]{a6a6ff}44.0 & \cellcolor[HTML]{ebebff}35.1 \\
             w/o ACID & 79.3 & 85.5 & 82.8 & 62.8 & 72.8 & 65.8 & 44.8 & 63.4 & 49.7 & 39.6 & 53.0 & 41.0 \\
             w/o SACE & 79.5 & 84.1 & 81.8 & 63.9 & 71.9 & 66.4 & \cellcolor[HTML]{ebebff}39.3 & 60.0 & 48.3 & 32.8 & 48.5 & 37.3 \\
             w/o SACE \& SESE & 79.1 & 80.7 & 78.5 & 63.3 & 71.7 & 64.2 & 43.4 & \cellcolor[HTML]{ebebff}59.3 & 46.9 & 37.3 & 50.0 & 39.6 \\
             \bottomrule[0.5mm]
        \end{tabular}
    }
    \caption{Ablation results of Track-SQL on SparC validation set by difficulty. The numbers in brackets () indicate the number of samples.}
    \label{tab:ablation_by_difficulty_sparc}
\end{table*}

\begin{table*}[]
    \centering
    \renewcommand\arraystretch{1.4}
    \resizebox{\textwidth}{!}{
        \begin{tabular}{c|ccc|ccc|ccc|ccc}
             \toprule[0.5mm]
             & \multicolumn{3}{c|}{Easy (417)} & \multicolumn{3}{c|}{Medium (320)} & \multicolumn{3}{c|}{Hard (163)} & \multicolumn{3}{c}{Extra (107)}\\
             & EM & EX & TS & EM & EX & TS & EM & EX & TS & EM & EX & TS \\
             \midrule
             Track-SQL & \textbf{80.6} & \textbf{85.4} & \textbf{82.5} & \textbf{51.6} & \textbf{67.5} & \textbf{55.6} & \textbf{36.2} & \textbf{59.5} & \textbf{46.0} & \textbf{24.3} & \textbf{39.3} & \textbf{28.0} \\
             w/o SESE & \cellcolor[HTML]{ccccff}75.1 & 79.4 & 76.7 & 50.0 & \cellcolor[HTML]{ccccff}60.9 & \cellcolor[HTML]{ebebff}51.6 & 36.2 & \cellcolor[HTML]{ccccff}53.4 & 46.0 & 25.2 & 38.3 & 31.8 \\
             w/o ACID & 78.2 & 83.5 & 81.1 & 52.8 & 65.0 & 56.2 & \cellcolor[HTML]{ebebff}31.3 & 55.8 & \cellcolor[HTML]{ebebff}44.2 & 22.4 & 37.4 & 26.2 \\
             w/o SACE & 78.2 & 83.5 & 80.3 & 50.0 & 61.6 & 51.9 & 35.6 & 62.0 & 48.5 & 29.0 & 43.9 & 36.4 \\
             w/o SACE \& SESE & 75.8 & \cellcolor[HTML]{a6a6ff}77.7 & \cellcolor[HTML]{ccccff}75.8 & \cellcolor[HTML]{ebebff}48.4 & \cellcolor[HTML]{ccccff}60.9 & \cellcolor[HTML]{ebebff}51.6 & 38.0 & 66.3 & 52.8 & \cellcolor[HTML]{a6a6ff}16.8 & \cellcolor[HTML]{ebebff}35.5 & \cellcolor[HTML]{ebebff}24.3 \\
             \bottomrule[0.5mm]
        \end{tabular}
    }
    \caption{Ablation results of Track-SQL on CoSQL validation set by difficulty. The numbers in brackets () indicate the number of samples.}
    \label{tab:ablation_by_difficulty_cosql}
\end{table*}

\subsubsection{Ablation by turn}
\label{sec:ablation_by_turn}
Table \ref{tab:ablation_by_turn_sparc} and Table \ref{tab:ablation_by_turn_cosql} present the ablation study results of Track-SQL under different numbers of interaction turns. A comprehensive analysis of the experimental results on the SparC and CoSQL datasets indicates that the SACE module continues to perform well in multi-turn interaction scenarios. Specifically, in the CoSQL dataset, when the number of interaction turns exceeds four, the SACE module achieves a performance gain of 2.8$\%$. In the SparC dataset, when the number of interaction turns reaches four, there is a 2.3$\%$ improvement. Additionally, the effectiveness of the ACID module and the SESE module is not affected by the number of interaction turns, and they consistently enhance the performance of the SQL generator across various samples.

\begin{table*}[]
    \centering
    \large
    \renewcommand\arraystretch{1.2}
    \resizebox{\textwidth}{!}{
        \begin{tabular}{c|ccc|ccc|ccc|ccc|ccc}
            \toprule[0.5mm]
            & \multicolumn{3}{c|}{Turn 1 (422)} & \multicolumn{3}{c|}{Turn 2 (422)} & \multicolumn{3}{c|}{Turn 3 (270)} & \multicolumn{3}{c|}{Turn 4 (88)} & \multicolumn{3}{c}{Turn > 4 (1)}\\
            & EM & EX & TS & EM & EX & TS & EM & EX & TS & EM & EX & TS & EM & EX & TS \\
            \midrule
            Track-SQL & \textbf{74.2} & \textbf{79.6} & \textbf{76.5} & \textbf{65.9} & \textbf{75.6} & \textbf{67.8} & \textbf{57.8} & \textbf{67.4} & \textbf{59.6} & \textbf{51.1} & \textbf{68.2} & \textbf{60.2} & \textbf{0} & \textbf{100} & \textbf{100} \\
            w/o SESE & \cellcolor[HTML]{ebebff}70.1 & \cellcolor[HTML]{ebebff}75.6 & \cellcolor[HTML]{ebebff}71.8 & \cellcolor[HTML]{ebebff}60.9 & \cellcolor[HTML]{ccccff}69.2 & \cellcolor[HTML]{ccccff}61.4 & \cellcolor[HTML]{ccccff}51.1 & \cellcolor[HTML]{ccccff}60.4 & \cellcolor[HTML]{ccccff}53.0 & \cellcolor[HTML]{a6a6ff}40.9 & \cellcolor[HTML]{a6a6ff}58.0 & \cellcolor[HTML]{ccccff}52.3 & 0 & 0 & 0  \\
            w/o ACID  & 72.5 & 80.6 & 77.3 & 64.2 & 75.1 & 66.6 & 56.7 & 67.8 & 59.3 & 54.5 & 64.8 & 56.8 & 0 & 0 & 0 \\
            w/o SACE & 72.3 & 79.1 & 75.8 & 63.0 & 72.3 & 64.9 & 55.9 & 65.9 & 59.6 & 51.1 & 65.9 & 60.2 & 0 & 0  & 0  \\
            w/o SACE \& SESE & 71.3 & 77.0 & 72.7 & 63.7 & 70.9 & 64.0 & 58.9 & 67.0 & 58.5 & 51.1 & 61.4 & 54.5 & 0 & 0 & 0 \\
             \bottomrule[0.5mm]
        \end{tabular}
    }
    \caption{Ablation results of Track-SQL on SparC validation set by turn. The numbers in brackets () indicate the number of samples.}
    \label{tab:ablation_by_turn_sparc}
\end{table*}

\begin{table*}[]
    \centering
    \renewcommand\arraystretch{1.2}
    \resizebox{\textwidth}{!}{
        \begin{tabular}{c|ccc|ccc|ccc|ccc|ccc}
             \toprule[0.5mm]
             & \multicolumn{3}{c|}{Turn 1(293)} & \multicolumn{3}{c|}{Turn 2 (285)} & \multicolumn{3}{c|}{Turn 3 (244)} & \multicolumn{3}{c|}{Turn 4 (114)} & \multicolumn{3}{c}{Turn > 4 (71)}\\
             & EM & EX & TS & EM & EX & TS & EM & EX & TS & EM & EX & TS & EM & EX & TS \\
             \midrule
             Track-SQL & \textbf{64.8} & \textbf{76.5} & \textbf{70.3} & \textbf{58.2} & \textbf{72.3} & \textbf{63.9} & \textbf{56.6} & \textbf{68.0} & \textbf{58.2} & \textbf{55.3} & \textbf{64.9} & \textbf{55.3} & \textbf{40.8} & \textbf{57.7} & \textbf{47.9} \\
             w/o SESE & \cellcolor[HTML]{ebebff}61.4 & \cellcolor[HTML]{ccccff}70.0 & \cellcolor[HTML]{ccccff}64.5 & 57.9 & 67.4 & 62.8 & \cellcolor[HTML]{ccccff}50.8 & \cellcolor[HTML]{ebebff}64.3 & \cellcolor[HTML]{ebebff}54.5 & 52.6 & \cellcolor[HTML]{a6a6ff}56.1 & 51.8 & 42.3 & \cellcolor[HTML]{ccccff}50.7 & 47.9 \\
             w/o ACID & 65.2 & 73.4 & 68.6 & 55.1 & 69.1 & 62.8 & 54.5 & 66.8 & 58.6 & 51.8 & 62.3 & 53.5 & 42.3 & 57.7 & 47.9 \\
             w/o SACE & 66.2 & 75.8 & 69.3 & \cellcolor[HTML]{ebebff}52.6 & \cellcolor[HTML]{ebebff}66.7 & \cellcolor[HTML]{ebebff}59.6 & 57.8 & 69.3 & 59.4 & 52.6 & 64.0 & 58.8 & 42.3 & 54.9 & 47.9 \\
             w/o SACE \& SESE & 62.8 & 71.0 & 65.2 & 55.1 & 68.1 & 62.5 & 52.5 & 65.2 & 55.7 & \cellcolor[HTML]{ebebff}50.0 & 58.8 & \cellcolor[HTML]{ebebff}50.0 & \cellcolor[HTML]{ebebff}35.2 & 52.1 & \cellcolor[HTML]{ebebff}43.7 \\
             \bottomrule[0.5mm]
        \end{tabular}
    }
    \caption{Ablation results of Track-SQL on CoSQL validation set by turn. The numbers in brackets () indicate the number of samples.}
    \label{tab:ablation_by_turn_cosql}
\end{table*}

\subsubsection{Ablation experiment performance under diverse language models}
\label{sec:ablation_by_differentLLMs}
\begin{figure}[h]
  \includegraphics[width=\linewidth]{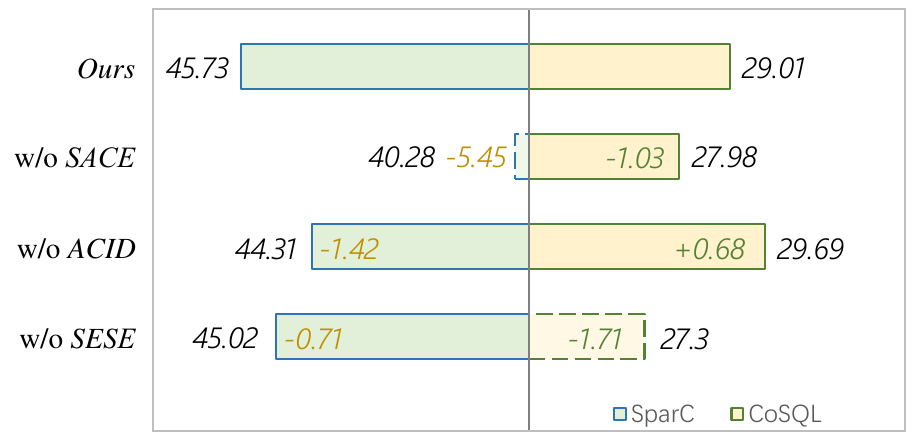} 
  \caption {The results of the ablation study on the Codellama 7B+Track-SQL model on the SparC and CoSQL dev sets (calculated using the multi-turn TS metrics).}
  \label{fig:codellama}
\end{figure}

\begin{figure}[h]
  \includegraphics[width=\linewidth]{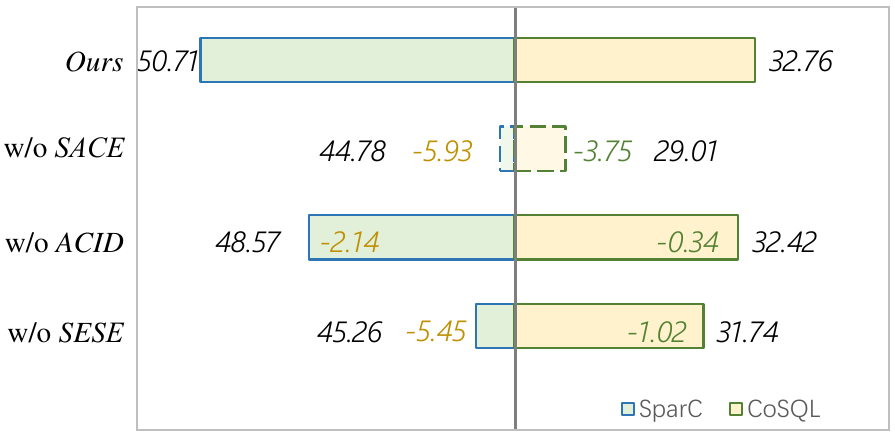}
  \caption {The results of the ablation study on the DeepSeek 7B+Track-SQL model on the SparC and CoSQL dev sets (calculated using the multi-turn TS metrics).}
  \label{fig:deepseek}
\end{figure}

\begin{figure}[h]
  \includegraphics[width=\linewidth]{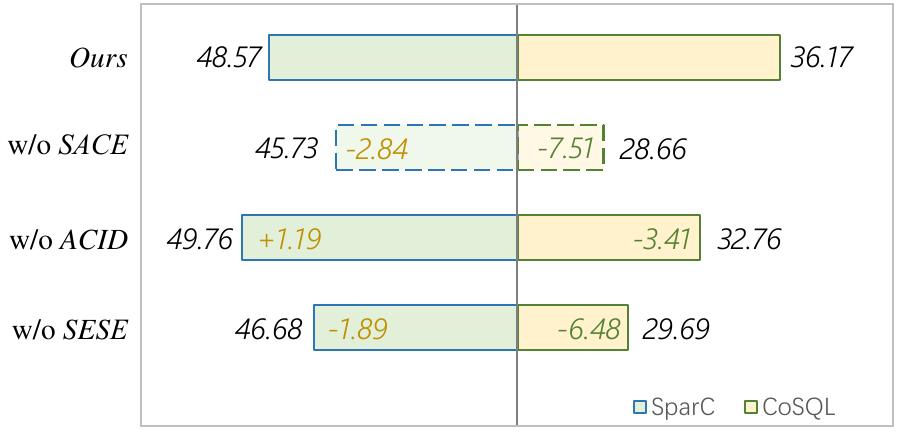}
  \caption {The results of the ablation study on the Mistral 7B+Track-SQL model on the SparC and CoSQL dev sets (calculated using the multi-turn TS metrics).}
  \label{fig:mistral}
\end{figure}

Figures \ref{fig:codellama}, \ref{fig:deepseek} and \ref{fig:mistral} present the ablation study results of the Track-SQL framework's modules on the Codellama, Mistral, and DeepSeek models. The data on the left and right sides of the charts correspond to the test results from the SparC and CoSQL development sets, respectively. The evaluation metric used is the TS score under the IM environment. The results show that on various 7B benchmark models, the components of the Track-SQL framework consistently demonstrate significant effectiveness, particularly the SESE and SACE modules, whose outstanding performance is especially noteworthy.

\subsection{Detailed evaluation of the Semantic-Enhanced Schema Extractor}
\label{sec:comprehensive_eval_schema_item_extractor}
A more comprehensive evaluation of the extractor is shown in Table \ref{tab:appendix_SESE_sparc} and \ref{tab:appendix_SESE_cosql}.
Tables \ref{tab:appendix_SESE_sparc} and \ref{tab:appendix_SESE_cosql} present the ablation study results of our proposed semantic-enhanced schema extractor on the SparC and CoSQL datasets, using overall classification accuracy as the evaluation metric. This evaluation method treats the classification probabilities of all columns as a unified whole, without distinguishing columns based on the table structure, nor imposing penalties for missing or redundant schemas. Therefore, while this measure of classification accuracy is somewhat rough, it still holds some reference value. The data from both tables indicate that the classification accuracy decreases to varying degrees when the semantic enhancement module is removed; further removal of the column enhancement components used in RESDSQL leads to an additional drop in accuracy. This suggests that the semantic enhancement module has a significant improvement effect on the base classifier and supports dynamic schema linking functionality.

\begin{table*}[h]
    \centering
    \large
    \renewcommand\arraystretch{1.45}
    \resizebox{\textwidth}{!}{
        \begin{tabular}{c|ccc|ccc|ccc|ccc}
            \toprule[0.5mm] 
            & \multicolumn{3}{c|}{\textbf{Table-Weighted Avg}} & \multicolumn{3}{c|}{\textbf{Table-Macro Avg}} & \multicolumn{3}{c|}{\textbf{Column-Weighted Avg}} & \multicolumn{3}{c}{\textbf{Column-Macro Avg}} \\
            & Precision & Recall & F1 & Precision & Recall & F1 & Precision & Recall & F1 & Precision & Recall & F1 \\
            \midrule
            SESE & 96.48 & 96.43 & 96.45 & 95.74 & \textbf{96.42} & 96.07 & \textbf{97.09} & \textbf{96.93} & \textbf{96.99} & 90.95 & \textbf{94.34} & \textbf{92.55} \\
            w/o com-enh & 96.18 & 96.11 & 96.12 & 95.32 & 96.16 & 95.72 & 96.91 & 96.68 & 96.76 & 90.11 & 94.24 & 92.03 \\
            w/o com-enh \& col-enh & \textbf{96.58} & \textbf{96.57} & \textbf{96.57} & \textbf{96.09} & 96.31 & \textbf{96.20} & 96.70 & 96.67 & 96.69 & \textbf{91.31} & 92.05 & 91.68 \\
            \bottomrule[0.5mm]
        \end{tabular}
    }
    \caption{The general precision index score of the semantic-enhanced schema extractor under the SparC dataset}
    \label{tab:appendix_SESE_sparc}
\end{table*}

\begin{table*}[h]
    \centering
    \large
    \renewcommand\arraystretch{1.45}
    \resizebox{\textwidth}{!}{
        \begin{tabular}{c|ccc|ccc|ccc|ccc}
            \toprule[0.5mm] 
            & \multicolumn{3}{c|}{\textbf{Table-Weighted Avg}} & \multicolumn{3}{c|}{\textbf{Table-Macro Avg}} & \multicolumn{3}{c|}{\textbf{Column-Weighted Avg}} & \multicolumn{3}{c}{\textbf{Column-Macro Avg}} \\
            & Precision & Recall & F1 & Precision & Recall & F1 & Precision & Recall & F1 & Precision & Recall & F1 \\
            \midrule
            SESE & \textbf{95.33} & \textbf{95.28} & \textbf{95.30} & \textbf{94.29} & \textbf{94.87} & \textbf{94.57} & \textbf{96.85} & 96.71 & 96.77 & 89.70 & \textbf{92.65} & \textbf{91.11} \\
            w/o com-enh & 95.32 & 95.30 & 95.31 & 94.47 & 94.66 & 94.57 & 96.81 & \textbf{96.79} & \textbf{96.80} & \textbf{90.90} & 91.24 & 91.07 \\
            w/o com-enh \& col-enh & 94.65 & 94.39 & 94.45 & 92.83 & 94.64 & 93.64 & 96.40 & 95.98 & 96.13 & 86.97 & 92.78 & 89.59 \\
            \bottomrule[0.5mm]
        \end{tabular}
    }
    \caption{The general precision index score of the semantic-enhanced schema extractor under the CoSQL dataset}
    \label{tab:appendix_SESE_cosql}
\end{table*}

\subsection{Case Study}
\label{sec:case study}
In ablation experiments conducted on the SparC dataset, we randomly selected several results generated by the model and will provide a detailed analysis of these samples in this section. All selected samples are listed in Table \ref{tab:case study}, and all examples are based on the deepseek-coder-6.7b model.

\noindent\textbf{w/o SACE Case Analysis }
In the first case, there is a notable correlation between question\#2 and question\#3; with the support of $SQL_{base}$, Track-SQL is able to generate valid SQL statements that include correct foreign key joins. Specifically, during the second turn of reasoning, the model successfully generates $SQL_{base}$, which simplifies the task of generating the foreign key join SQL statement in the third turn. However, when the SACE module is removed, this cumulative effect no longer exists, leading to generated SQL statements that fail to execute the correct foreign key join operations. In the second case, when Track-SQL enters the reasoning phase for the third question, it leverages the $SQL_{base}$ generated from the first question to achieve precise predictions of schemas.

\noindent\textbf{w/o ACID Case Analysis }
From the third and fourth examples, it can be observed that removing the ACID module diminishes the model's ability to identify whether the user needs to display all columns of a table, resulting in errors in the generation of SQL statements.

\noindent\textbf{w/o SESE Case Analysis }
The fifth and sixth examples demonstrate that the removal of the SESE module leads to a decrease in accuracy when generating schemas. Specifically, in the fifth example, the model-generated result lacks the "\textit{Continent}" column; whereas in the sixth example, the model incorrectly generates the "\textit{tv\_channel.series\_name}" column. In contrast, the $SQL_{base}$ model reduces instances of erroneous generation and missing columns.

\begin{table*}[]
    \centering
    \large
    \resizebox{\textwidth}{!}{
    \begin{tabular}{l|c|l}
        \toprule[0.5mm]
        \specialrule{0em}{2.5pt}{2.5pt}
        \makebox[0.10\textwidth][l]{\multirow{10}{*}{Case\#1}} & \makebox[0.15\textwidth][c]{\makecell[c]{Question\#1\\Question\#2\\Question\#3}} & \makebox[0.75\textwidth][l]{\makecell[l]{What is every student's id? \\ Of those ids, which correspond to those who own cats as pets? \\ List all the other ids.}} \\
        \specialrule{0em}{2.5pt}{2.5pt}
        \cdashline{2-3}
        \specialrule{0em}{2.5pt}{2.5pt}
        & Gold & \textbf{\makecell[l]{SELECT stuid FROM student EXCEPT SELECT T1.stuid FROM student AS T1 JOIN \\has\_pet AS T2 ON T1.stuid  =  T2.stuid JOIN pets AS T3 ON T3.petid  =  T2.petid WHERE \\T3.pettype  =  'cat'}} \\
        \specialrule{0em}{2.5pt}{2.5pt}
        \cdashline{2-3}
        \specialrule{0em}{2.5pt}{2.5pt}
        & w/o SACE & \makecell[l]{SELECT stuid FROM has\_pet except SELECT stuid FROM has\_pet JOIN pets ON has\_pet.petid\\ = pets.petid WHERE pettype = 'cat'} \\
        \specialrule{0em}{2.5pt}{2.5pt}
        \cdashline{2-3}
        \specialrule{0em}{2.5pt}{2.5pt}
        & Track-SQL & \makecell[l]{SELECT stuid FROM student except \textcolor[HTML]{006633}{SELECT student.stuid FROM student JOIN has\_pet ON}\\ \textcolor[HTML]{006633}{student.stuid = has\_pet.stuid JOIN pets ON has\_pet.petid = pets.petid WHERE pets.pettype = 'cat'}} \\
        \specialrule{0em}{2.5pt}{2.5pt}
        \cdashline{2-3}
        \specialrule{0em}{2.5pt}{2.5pt}
        & $SQL_{base}$ & \makecell[l]{\textcolor[HTML]{006633}{SELECT student.stuid FROM student JOIN has\_pet ON student.stuid = has\_pet.stuid JOIN pets}\\ \textcolor[HTML]{006633}{ON has\_pet.petid = pets.petid WHERE pets.pettype = 'cat'}} \\
        \specialrule{0em}{2.5pt}{2.5pt}
        \cline{1-3}
        \specialrule{0em}{2.5pt}{2.5pt}
        \multirow{10}{*}{Case\#2}& \makecell[c]{Question\#1\\Question\#2\\Question\#3} & \makecell[l]{How many car models are produced in total? \\ How many in Germany? \\ How about in Japan?} \\
        \specialrule{0em}{2.5pt}{2.5pt}
        \cdashline{2-3}
        \specialrule{0em}{2.5pt}{2.5pt}
        & Gold & \textbf{\makecell[l]{SELECT count(*) FROM MODEL\_LIST AS T1 JOIN CAR\_MAKERS AS T2 ON \\T1.Maker = T2.Id JOIN COUNTRIES AS T3 ON T2.Country  =  T3.CountryId WHERE\\ T3.CountryName =  'japan'}} \\
        \specialrule{0em}{2.5pt}{2.5pt}
        \cdashline{2-3}
        \specialrule{0em}{2.5pt}{2.5pt}
        & w/o SACE & \makecell[l]{SELECT COUNT( model\_list.model ) FROM model\_list JOIN car\_names ON model\_list.model =\\ car\_names.model JOIN countries ON car\_names.makeid = countries.countryid WHERE \\countries.countryname = 'japan'} \\
        \specialrule{0em}{2.5pt}{2.5pt}
        \cdashline{2-3}
        \specialrule{0em}{2.5pt}{2.5pt}
        & Track-SQL & \makecell[l]{\textcolor[HTML]{006633}{SELECT COUNT( * ) FROM model\_list} JOIN car\_makers ON model\_list.maker = car\_makers.id \\JOIN countries ON car\_makers.country = countries.countryid WHERE countries.countryname = \\'japan'} \\
        \specialrule{0em}{2.5pt}{2.5pt}
        \cdashline{2-3}
        \specialrule{0em}{2.5pt}{2.5pt}
        & $SQL_{base}$ & \makecell[l]{\textcolor[HTML]{006633}{SELECT COUNT( * ) FROM model\_list}} \\
        \specialrule{0em}{2.5pt}{2.5pt}
        \cline{1-3}
        \midrule[1pt]
        \specialrule{0em}{2.5pt}{2.5pt}
        \multirow{6}{*}{Case\#3}& Question & What are all the nations? \\
        \specialrule{0em}{2.5pt}{2.5pt}
        \cdashline{2-3}
        \specialrule{0em}{2.5pt}{2.5pt}
        & Gold & \textbf{SELECT * FROM country} \\
        \specialrule{0em}{2.5pt}{2.5pt}
        \cdashline{2-3}
        \specialrule{0em}{2.5pt}{2.5pt}
        & w/o ACID & SELECT \textbf{\textcolor{red}{name}} FROM country \\
        \specialrule{0em}{2.5pt}{2.5pt}
        \cdashline{2-3}
        \specialrule{0em}{2.5pt}{2.5pt}
        & Track-SQL & SELECT \textcolor[HTML]{006633}{*} FROM country \\
        \specialrule{0em}{2.5pt}{2.5pt}
        \cline{1-3}
        \specialrule{0em}{2.5pt}{2.5pt}
        \multirow{6}{*}{Case\#4}& Question & Show all the available features. \\
        \specialrule{0em}{2.5pt}{2.5pt}
        \cdashline{2-3}
        \specialrule{0em}{2.5pt}{2.5pt}
        & Gold & \textbf{SELECT * FROM Other\_Available\_Features} \\
        \specialrule{0em}{2.5pt}{2.5pt}
        \cdashline{2-3}
        \specialrule{0em}{2.5pt}{2.5pt}
        & w/o ACID & SELECT \textbf{\textcolor{red}{feature\_name}} FROM other\_available\_features \\
        \specialrule{0em}{2.5pt}{2.5pt}
        \cdashline{2-3}
        \specialrule{0em}{2.5pt}{2.5pt}
        & Track-SQL & SELECT \textcolor[HTML]{006633}{*} FROM other\_available\_features \\
        \specialrule{0em}{2.5pt}{2.5pt}
        \cline{1-3}
        \midrule[1pt]
        \specialrule{0em}{2.5pt}{2.5pt}
        \multirow{6}{*}{Case\#5}& Question & Which continents have an average life expectancy less than age 72?\\
        \specialrule{0em}{2.5pt}{2.5pt}
        \cdashline{2-3}
        \specialrule{0em}{2.5pt}{2.5pt}
        & Gold & \textbf{\makecell[l]{SELECT sum(Population) ,  avg(LifeExpectancy),  Continent FROM country GROUP BY\\ Continent HAVING avg(LifeExpectancy)  <  72}} \\
        \specialrule{0em}{2.5pt}{2.5pt}
        \cdashline{2-3}
        \specialrule{0em}{2.5pt}{2.5pt}
        & w/o SESE & \makecell[l]{SELECT \textbf{\textcolor{red}{avg(lifeexpectancy), sum(population)}} FROM country GROUP BY continent HAVING \\avg(lifeexpectancy) < 72\\} \\
        \specialrule{0em}{2.5pt}{2.5pt}
        \cdashline{2-3}
        \specialrule{0em}{2.5pt}{2.5pt}
        & Track-SQL & \makecell[l]{SELECT avg(lifeexpectancy), sum(population), \textcolor[HTML]{006633}{continent} FROM country GROUP BY continent\\ HAVING avg(lifeexpectancy) < 72} \\
        \specialrule{0em}{2.5pt}{2.5pt}
        \cline{1-3}
        \specialrule{0em}{2.5pt}{2.5pt}
        \multirow{6}{*}{Case\#6}& \makecell[c]{Question\#1 \\ Question\#2} & \makecell[l]{Tell me the director of the cartoon named "Day of the Dark Knight!". \\ What is the channel of this cartoon?} \\
        \specialrule{0em}{2.5pt}{2.5pt}
        \cdashline{2-3}
        \specialrule{0em}{2.5pt}{2.5pt}
        & Gold & \textbf{SELECT Channel FROM Cartoon WHERE Title = "Day of the Dark Knight!"} \\
        \specialrule{0em}{2.5pt}{2.5pt}
        \cdashline{2-3}
        \specialrule{0em}{2.5pt}{2.5pt}
        & w/o SESE & \makecell[l]{SELECT \textbf{\textcolor{red}{tv\_channel.series\_name}} FROM \textbf{\textcolor{red}{cartoon JOIN tv\_channel ON cartoon.channel =}} \\\textbf{\textcolor{red}{tv\_channel.id}} WHERE cartoon.title = 'Day of the Dark Knight!'} \\
        \specialrule{0em}{2.5pt}{2.5pt}
        \cdashline{2-3}
        \specialrule{0em}{2.5pt}{2.5pt}
        & Track-SQL & SELECT \textcolor[HTML]{006633}{channel} FROM cartoon WHERE Title = 'Day of the Dark Knight!' \\
        \bottomrule[0.5mm]
    \end{tabular}}
    \caption{Case analysis of ablation experiments on the SParC dataset}
    \label{tab:case study}
\end{table*}

\subsection{Resource overhead experiment}
\label{sec:resource_overhead_experiment}

\subsubsection{Quantitative analysis of time overhead}
\label{sec:quantitative_analysis_of_time}
In this section, we conducted detailed experiments on Track-SQL in the time dimension. As shown in Table \ref{tab:appendix_inference_time}. This includes the average inference time and training time of the schema extractor and the SQL generator, and all experiments were carried out on the same hardware specified in the main text.

Specifically, the average inference time for processing each sample using the Track-SQL framework is 1.352 seconds, which is significantly faster than context learning-based methods like CoE-SQL, which requires an average of 3.544 seconds. This highlights the obvious efficiency advantage of Track-SQL inference. In addition, when the batch size is 6, the schema extractor and the SQL generator reach the best model state within a reasonable training duration, as shown in Table 6. It is worth noting that the inference time was measured on the validation set, while the training time was based on the SparC and CoSQL training sets, which include 9,025 and 7,343 entries respectively.

\begin{table*}[h]
    \centering
    \large
    \renewcommand\arraystretch{1.45}
    \resizebox{\textwidth}{!}{
        \begin{tabular}{c|cc|cc|cc|cc}
            \toprule[0.5mm] 
            & \multicolumn{2}{c|}{\textbf{SESE(Inference)}} & \multicolumn{2}{c|}{\textbf{SQL Generator(Inference)}} & \multicolumn{2}{c|}{\textbf{SESE(Train)}} & \multicolumn{2}{c}{\textbf{SQL Generator(Train)}} \\
            & Total time(s) & Avg Time(s) & Total time(s) & Avg Time(s)
            & Total time(h) & Avg Time(s) & Total time(h) & Avg Time(s)\\
            \midrule
            SparC  & 240.348±1.45 & 0.20±0.00013 & 1386.98±2.13 & 1.152±0.025
                   & 30.93h±2.78 & 3.88±0.43 & 1.45±0.21 & 11.23±2.31 \\
            CoSQL  & 214.456±2.56 & 0.21±0.0012 & 1170.41±1.56 & 1.15±0.012
                    & 27.49h±2.43 & 3.92±0.29 & 1.61±0.33 & 11.43±2.51\\
            \bottomrule[0.5mm]
        \end{tabular}
    }
    \caption{Inference time performance of the Track-SQL framework. Avg Time(s) in terms of training represents the time consumption of a single batch, and Total time(h) represents the time required to obtain the best model. The SQL generator is based on Deepseek 7B}
    \label{tab:appendix_inference_time}
\end{table*}

\subsubsection{Memory Costs of Training and Inference}
\label{sec:memory_costs}
As shown in Table \ref{tab:appendix_memory_costs}, the Track-SQL framework is fully capable of performing inference on a 24G graphics card, demonstrating its characteristic of maintaining low cost while maintaining high accuracy. In terms of model training, by reducing the batch size, Track-SQL can also perform training tasks on low-capacity graphics cards.

\begin{table*}[h]
    \centering
    \large
    \renewcommand\arraystretch{1.45}
    \resizebox{\textwidth}{!}{
        \begin{tabular}{c|c|c|c|c}
            \toprule[0.5mm] 
            & \textbf{SESE(Inference)} & \textbf{SQL Generator(Inference)} 
            & \textbf{SESE(Train)} & \textbf{SQL Generator(Train)}
            \\
            \midrule
            Graphics Memory(GB) & 2.235 & 16.477 
                            & 20.997 & 62.194 \\
            \bottomrule[0.5mm]
        \end{tabular}
    }
    \caption{Memory Costs of Training and Inference in the Track-SQL Framework. In the training stage, the batch size is set to 6, and in the inference stage, the batch size is set to 1.The SQL generator is based on Deepseek 7B}
    \label{tab:appendix_memory_costs}
\end{table*}

\section{Appendix B}
\label{sec:appendix b}
In this section, we provide additional details about the methods described in the paper. \\
- In Section \ref{sec:loss function}, we describe the loss function used for the Semantic-enhanced Schema Extractor. We adopted the focal loss used by REDSQL.\\
- In Section \ref{sec:annotation_prompt}, we describe the prompts used to generate annotations for database schemas, which serve as input to the GPT3.5-turbo model.\\
- In Section \ref{sec:Format for fine-tuning SQL generation}, we detail the input and output formats used for supervised fine-tuning of the SQL expert model.

\subsection{Loss Function}
\label{sec:loss function}
Since an SQL query typically involves only a few tables and columns from the database, the label distribution in the training set is highly imbalanced. As a result, the number of negative samples is often several times greater than the number of positive samples, which can lead to significant training bias. To alleviate this issue, we adopt focal loss \cite{ross2017focal} as our classification loss. Then, we formulate the loss function for the multi-turn schema extractor using a multi-task learning approach, where the loss function consists of table classification loss and column classification loss:
\begin{equation}
\begin{split}
    L_{1}=\frac{1}{N}\sum_{i=1}^{N}FL(y_{i},\hat{y}_{i})+\\\frac{1}{M}\sum_{i=1}^{N} \sum_{k=1}^{n_{i}}FL(y_{i,k},\hat{y}_{i,k})
\end{split}
\end{equation}
where the focal loss function is denoted by $ FL $ and $ y_{i} $ represents the true label of the $ i^{th} $ table. $ y_{i} = 1 $ indicates that the table is referenced by the SQL, otherwise, it is 0. $y_{i,k}$ represents the true label of the $k^{th}$ column in the $i^{th}$ table. Similarly, $y_{i,k} = 1$ indicates that the column is referenced by an SQL, otherwise it is 0. $\hat{y}_{i}$ and $\hat{y}_{i,k}$ are predicted probabilities, which are estimated based on the table embeddings and column embeddings $T_{i}$ and $C_{i,k}$ through two different MLP modules:

\begin{equation}
    \hat{y}_{i}=\sigma ((\hat{T}_{i}U_{1}^{t}+b_{1}^{t})U_{2}^{t}+b_{2}^{t})
\end{equation}

\begin{equation}
    \hat{y}_{i,k}=\sigma ((C_{i,k}U_{i}^{c}+b_{1}^{c})U_{2}^{c}+b_{2}^{c})
\end{equation}
among them, $ U_{1}^{t}, U_{1}^{c} \in \mathbb{R}^{d \times \omega} $, $ b_{1}^{t}, b_{1}^{c} \in \mathbb{R}^{\omega} $, $ U_{2}^{t}, U_{2}^{c} \in \mathbb{R}^{2 \times \omega} $, $ b_{2}^{t}, b_{2}^{c} \in \mathbb{R}^{2} $ are trainable parameters, and $ \sigma(\cdot) $ denotes the Softmax function.

\subsection{Database schema annotation generation}
\label{sec:annotation_prompt}

Table \ref{tab:prompt_gpt} above displays the LLM input sequences used for generating annotations for database columns. These sequences include prompt statements, corresponding table names, column names, column data types, and several example values randomly drawn from the database. Additionally, Table \ref{tab:prompt_gpt} below lists the LLM input format used for generating annotations for entire database tables, which includes prompt statements, target table names, and their respective column names.

\begin{table*}[]
    \centering
    \resizebox{\textwidth}{!}{
    \begin{tabular}{|p{\textwidth}|}
        \hline
        \rule{0pt}{13pt}
        \centerline{\large{\textbf{Input for generating annotated descriptions of database columns}}}\\
        \hline
        You are a database schema designer who specializes in generating concise descriptions for table columns based on the provided column names, types, and sample values. The descriptions should be brief, adjective-noun phrases that reflect the nature of the data in the column.\\\\
        Table Name: \{table\_name\}\\
        Column: \{column\_name\}\\
        Type: \{column\_type\}\\
        Sample Values: \{values\}\\
        \hline
        \hline
        \rule{0pt}{13pt}
        \centerline{\large{\textbf{Input for generating annotated descriptions of database tables}}}\\
        \hline
        You are a database schema expert who is skilled at generating concise descriptions for database tables based on their names and column details. Your job is to create a brief, adjective-noun form description that captures the essence of the table. The description should be short and to the point, not exceeding a few words. \\ 
        You will be given the table name, column names, types, and sample values. \\
        Generate a descriptive phrase using this information. 
        The table name is `\{table\_name\}'. 
        It has the following columns: \{column\_strs\}\\\\
        Based on this information, please generate a concise description for the table.\\
        \hline
    \end{tabular}}
    \caption{This table lists LLM input sequences used for generating annotations for database columns and tables}
    \label{tab:prompt_gpt}
\end{table*}


\subsection{Format for fine-tuning SQL generation}
\label{sec:Format for fine-tuning SQL generation}
Table \ref{tab:prompt_finetune} shows the input and output formats used for supervised fine-tuning of the SQL generation model. Here, $E(\mathcal{S})$ represents the refined schema term sequence obtained through Semantic-enhanced Schema Extractor, $SQL_{base}$ is the historical reasoning SQL selected by Schema-aware Context Extractor, and $\mathcal{Q}_{ \leq m}$ is the combined sequence of historical questions and the current question. The output $s_{m}$ is the target SQL.

\begin{table*}[]
    \centering
    \resizebox{\textwidth}{!}{
    \begin{tabular}{|p{\textwidth}|}
        \hline
        \rule{0pt}{13pt}
        \centerline{\large{\textbf{Format for fine-tuning SQL generation}}}\\
        \hline
        \textbf{Input}: \\
        You are a SQL query generator that converts multi-turn questions along with associated database schema information into corresponding SQL statements. The multi-turn questions will be concatenated using the `$\&$' symbol, and you should generate the SQL statement that answers the current turn of the question based on the provided database schema.\\\\
        Each database schema is provided in the following format:\\
        Table name : Column name1, Column name2 Different tables are separated by the `|' symbol, and the order of table names and column names is relevant to the current question; those appearing earlier are more closely related.\\
        Base SQL: \{$SQL_{base}$\}\\
        database schema: \{$E(\mathcal{S})$\}\\
        question: \{$\mathcal{Q}_{\le m }$\}\\\\
        \textbf{Output}:\\
        ``` \\
        \{$s_{m}$\} ;\\
        ''' \\
        < | end\_of\_sentence | > \\
        \hline
    \end{tabular}}
    \caption{Format for fine-tuning SQL generation}
    \label{tab:prompt_finetune}
\end{table*}

\end{document}